\documentclass[runningheads]{llncs}
\usepackage{graphicx}
\usepackage{tikz}
\usepackage{comment}
\usepackage{amsmath,amssymb} 
\usepackage{color}
\usepackage{floatrow}

\usepackage{adjustbox}
\usepackage{tabularx}
\usepackage{booktabs}
\usepackage{multirow}
\usepackage[font=footnotesize,skip=0pt]{caption}

\usepackage{subcaption}

\usepackage{bbm}

\usepackage{caption}

\definecolor{OliveGreen}{rgb}{0,0.6,0}
\definecolor{SoftRed}{rgb}{1,0.2,0.2}

\usepackage{hyperref}
\hypersetup{
     colorlinks   = true,
     citecolor    = OliveGreen,
     linkcolor    = SoftRed
}
\usepackage[hyperpageref]{backref}
\begin{document}
\pagestyle{headings}
\mainmatter
\def\ECCVSubNumber{90}  

\title{PSUMNet: Unified Modality Part Streams are All You Need for Efficient Pose-based Action Recognition} 


\titlerunning{PSUMNet}
%
\author{Neel Trivedi 
\and Ravi Kiran Sarvadevabhatla}
\authorrunning{N. Trivedi et al.}
%
\institute{Centre for Visual Information Technology\\ IIIT Hyderabad, INDIA-500032
\email{\{neel.trivedi@research.,ravi.kiran@\}iiit.ac.in}}

\maketitle

\begin{center}
    \centering
    \includegraphics[width=0.9\textwidth]{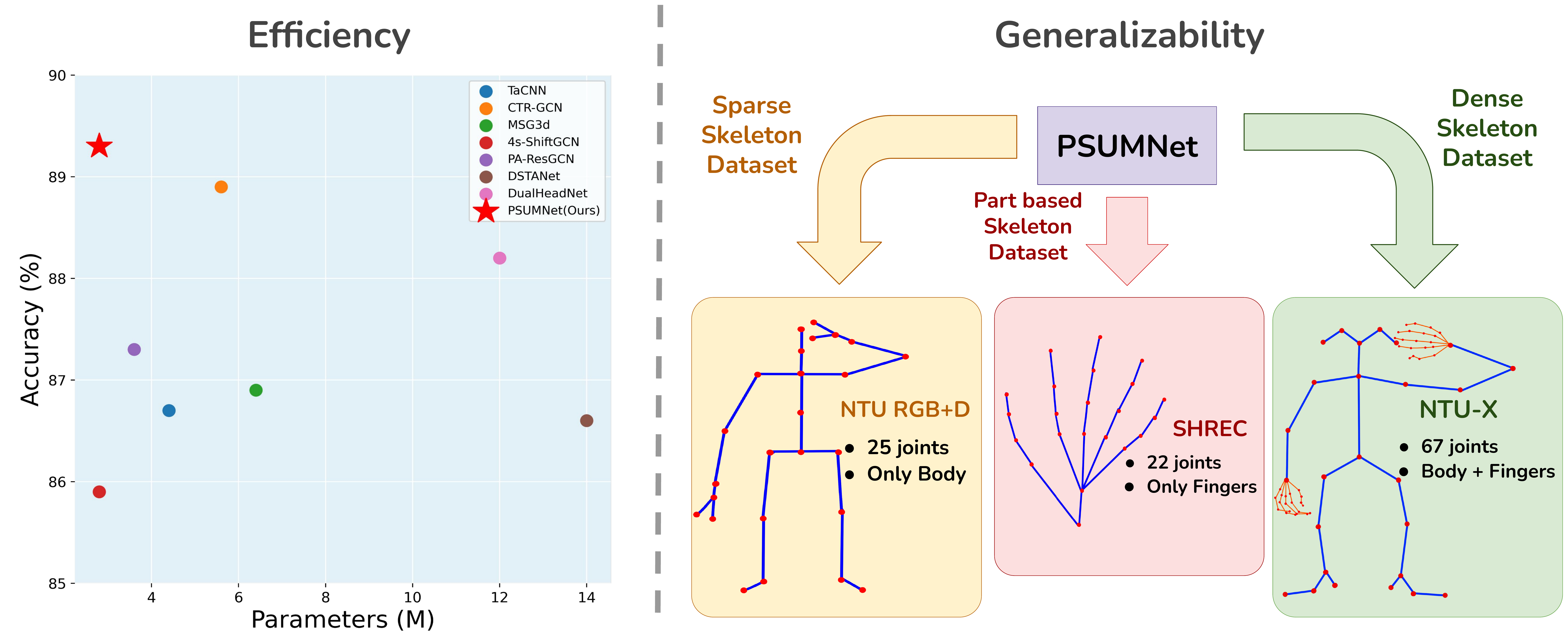}
  \captionof{figure}{The plot on left shows accuracy against \# parameters for our proposed architecture PSUMNet ($\color{red}{\star}$) and existing approaches for the large-scale NTURGB+D 120 human actions dataset (cross subject). PSUMNet achieves state of the art performance while competing recent methods use 100\%-400\% more parameters. The diagram on right illustrates that PSUMNet scales to sparse pose (SHREC~\cite{de20173d}) and dense pose (NTU-X~\cite{trivedi2021ntux}) configurations in addition to the popular NTURGB+D\cite{Liu_2019_NTURGBD120} configuration.}
  \label{fig:teaser}
\end{center}

\begin{abstract}
Pose-based action recognition is predominantly tackled by approaches which treat the input skeleton in a monolithic fashion, i.e. joints in the pose tree are processed as a whole. However, such approaches ignore the fact that action categories are often characterized by localized action dynamics involving only small subsets of part joint groups involving hands (e.g. `Thumbs up') or legs (e.g. `Kicking'). Although part-grouping based approaches exist, each part group is not considered within the global pose frame, causing such methods to fall short. Further, conventional approaches employ independent modality streams (e.g. joint, bone, joint velocity, bone velocity) and train their network multiple times on these streams, which massively increases the number of training parameters. To address these issues, we introduce PSUMNet, a novel approach for scalable and efficient pose-based action recognition. At the representation level, we propose a global frame based part stream approach as opposed to conventional modality based streams. Within each part stream, the associated data from multiple modalities is unified and consumed by the processing pipeline. Experimentally, PSUMNet achieves state of the art performance on the widely used NTURGB+D 60/120 dataset and dense joint skeleton dataset NTU 60-X/120-X. PSUMNet is highly efficient and outperforms competing methods which use 100\%-400\% more parameters. PSUMNet also generalizes to the SHREC hand gesture dataset with competitive performance. Overall, PSUMNet's scalability, performance and efficiency makes it an attractive choice for action recognition and for deployment on compute-restricted embedded and edge devices. Code and pretrained models can be accessed at \href{https://github.com/skelemoa/psumnet}{https://github.com/skelemoa/psumnet}.
\end{abstract}

\section{Introduction}
\label{sec:intro}

\begin{figure*}[]
  \centering
  \includegraphics[width=0.9\textwidth]{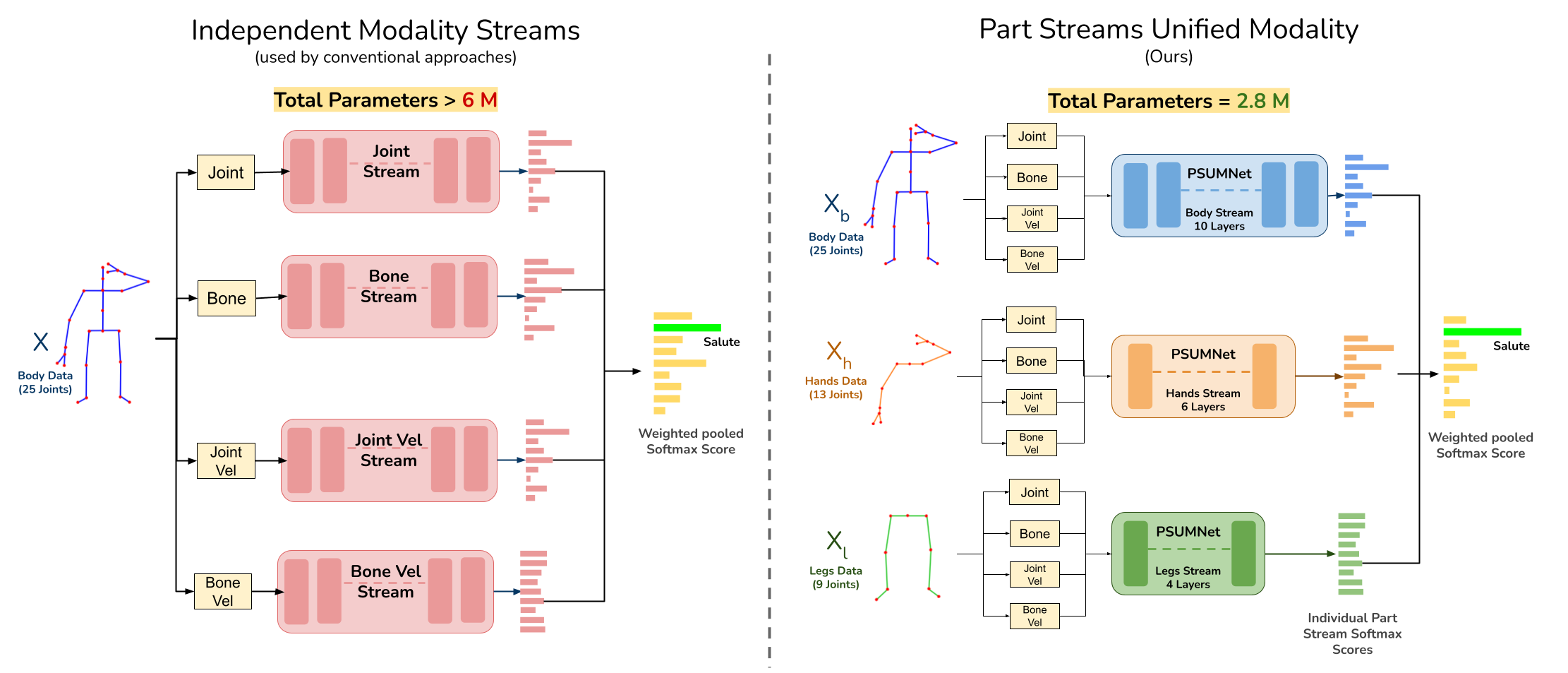}
  \caption{Comparison between conventional training procedure used in most of the previous approaches (left) and our approach (right). Conventional methods~\cite{chen2021channel,liu2020disentangling} use dedicated independent streams and train separate instances of the same network for each of the four modalities, i.e joint, bone, joint velocity and bone velocity. This method increases the number of total parameters by a huge margin and involves a monolithic representation. Our method processes the modalities in a unified manner and creates part group based independent stream with a superior performance compared to existing methods which use 100\%-400\% more parameters - see Fig.~\ref{fig:architecture} for architectural details of PSUMNet.}
  \label{fig:pipeline}
\end{figure*}


Skeleton based human action recognition at scale has gained a lot of focus recently, especially with the release of large scale skeleton action datasets such as NTURGB+D~\cite{Shahroudy_2016_CVPR} and NTURGB+D 120~\cite{Liu_2019_NTURGBD120}. A plethora of RNN~\cite{7298714,HAN201785}, CNN~\cite{zhang2019view,HernandezRuiz:2017:CDM:3123266.3123299} and GCN~\cite{stgcn2018aaai,liu2020disentangling} based approaches have been proposed to tackle this important problem. The success of approaches such as ST-GCN~\cite{stgcn2018aaai} which modeled spatio-temporal joint dynamics using GCN has given much prominence to GCN-based approaches. Furthermore, approaches such as RA-GCN~\cite{song2020richly} and ~2s-AGCN\cite{shi2019two} built upon this success and demonstrated additional gains by introducing multi modal (bone and velocity) streams -- see Fig.~\ref{fig:pipeline} (left). This multi stream approach has been adopted as convention by state of the art approaches. 

However, the conventional setup has three major drawbacks. \textit{First,} each modality stream is trained independently and the results are combined using late (decision) fusion. This deprives the processing pipeline from taking advantage of correlations across modalities. \textit{Second,} with addition of each new modality, the number of parameters increase by a significant margin since a separate network with the same model architecture is trained for each modality.
\textit{Third,} the skeleton is considered in a monolithic fashion. In other words,  the entire input pose tree at each time step is treated as a whole and at once. This is counter intuitive to the fact that a lot of action categories often involve only a subset of the available joints. For example, action categories such as ``Cutting paper" or ``Writing" can be easily identified using only hand joints whereas action categories such as ``Walking" or ``Kicking" can be easily identified using only leg joints. Additionally, monolithic processing increases compute requirements when the number of joints in the pose representation increases~\cite{trivedi2021ntux}. Non-monolithic approaches which decompose the pose tree into disjoint part groups do exist~\cite{thakkar2018partbased,song2020stronger}. However, each part group is not considered within the global pose frame, causing such methods to fall short. 

Our proposed approach tackles all of the aforementioned drawbacks - see Fig.~\ref{fig:pipeline} (right). Our contributions are the following:

\begin{itemize}
    \item We propose a unified modality processing approach as opposed to conventional independent modality approaches. This enables a significant reduction in the number of parameters. (Sec.~\ref{sec:psumnet})
    \item We propose a part based stream processing approach which enables richer and  dedicated representations for actions involving a subset of joints (Sec.~\ref{sec:partstream}). The part stream approach also enables efficient generalization to dense joint (NTU-X~\cite{trivedi2021ntux}) and small joint (SHREC~\cite{de20173d}) datasets. 
    \item Our architecture, dubbed Part Stream Unified Modality Network (PSUMNet)  achieves SOTA performance on NTU 60-X/120-X, and NTURGB+D 60/120 datasets compared to existing competing methods which use 100\%-400\% more parameters. PSUMNet also generalizes to SHREC hand gestures dataset with competitive performance. (Sec.~\ref{sec:results})
    \item We perform extensive experiments and ablations to analyze and demonstrate the superiority of PSUMNet. (Sec.~\ref{sec:experiments}).
\end{itemize}

The high accuracy provided by PSUMNet, coupled with its efficiency in terms of compute (number of parameters and floating-point operations) makes our approach an attractive choice for real world deployment on compute restricted embedded and edge devices - see Fig.~\ref{fig:teaser}. Code and pretrained models can be accessed at \href{https://github.com/skelemoa/psumnet}{https://github.com/skelemoa/psumnet}.

\section{Related Work}
\label{sec:related_work}
\noindent \textbf{Skeleton action recognition:} Since the release of large scale skeleton based datasets~\cite{Shahroudy_2016_CVPR,Liu_2019_NTURGBD120} various CNN \cite{zhang2019view,HernandezRuiz:2017:CDM:3123266.3123299,10.5555/3304415.3304527}, RNN \cite{7298714,HAN201785,zhang2019view,Zhao_2019_ICCV} and recently GCN based methods have been proposed for skeleton action recognition. ST-GCN~\cite{stgcn2018aaai} was the first successful approach to model the spatio-temporal relationships for skeleton actions at scale. Many state of the art approaches~\cite{cheng2020shiftgcn,liu2020disentangling,dstanet_accv2020,chen2021channel} have adopted and modified this approach to achieve superior results. However, these approaches predominantly process the skeleton joints in a monolithic manner, i.e these approaches process the entire input skeleton at once which can create a bottleneck when the input skeleton becomes denser, e.g. NTU-X~\cite{trivedi2021ntux}.

\noindent \textbf{Part based approaches:} The idea of grouping skeleton joints into different groups has few precedents. Du et al.~\cite{7298714} propose a RNN-based hierarchical grouping of part group representations. Thakkar et al.~\cite{thakkar2018partbased} propose a GCN based approach which applies  modified graph convolutions to different part groups. Huang et al.~\cite{Huang_Huang_Ouyang_Wang_2020} propose a GCN-based approach in which they utilize the higher order part level graph for better pooling and aligning of nodes in the main skeleton graph.  More recently, Song et al.~\cite{song2020stronger} propose a part-aware GCN method which utilizes part factorization to aid an attention mechanism to find the most informative part. Some previous part based approaches segment the limbs based on left and right orientation as well (left/right arm, left/right leg etc.)~\cite{song2020stronger,thakkar2018partbased}. Such segmentation leads to disjoint part groups which contain very small number of joints and are unable to convey useful information. In contrast, our part stream approach creates overlapping part groups with sufficient number of joints to model useful relationships. Also, each individual part group in our setup is registered to the global frame unlike the per-group coordinate system setup in existing approaches. In addition, we employ a combination of part group and coarse version of the full skeleton instead of part-group only approach seen in existing approaches. Our part stream approach allows each part based sub-skeleton to contribute towards the final prediction via decision fusion. To the best of our knowledge, such globally registered independent part stream approach has never been used before.

\noindent \textbf{Multi stream Training:} Earlier approaches~\cite{shi2019two,cheng2020shiftgcn,liu2020disentangling} and more recent approaches~\cite{chen2021channel,xu2021topology} create multiple modalities termed joint, bone and velocity from the raw input skeleton data. The conventional method is to train the given architecture multiple times using different modality data followed by decision fusion. However, this conventional approach with multiple versions of the base architecture greatly increases the total number of parameters. Song et al.~\cite{song2020stronger} attempt a unified modality pipeline wherein early fusion of different modality streams is used to achieve a unified modality representation. However, before the fusion, each modality is processed via multiple independent networks which again increases the count of trainable parameters.

\section{Methodology}
\label{sec:methodology}

\begin{figure*}[!t]
  \centering
  \includegraphics[width=0.8\textwidth]{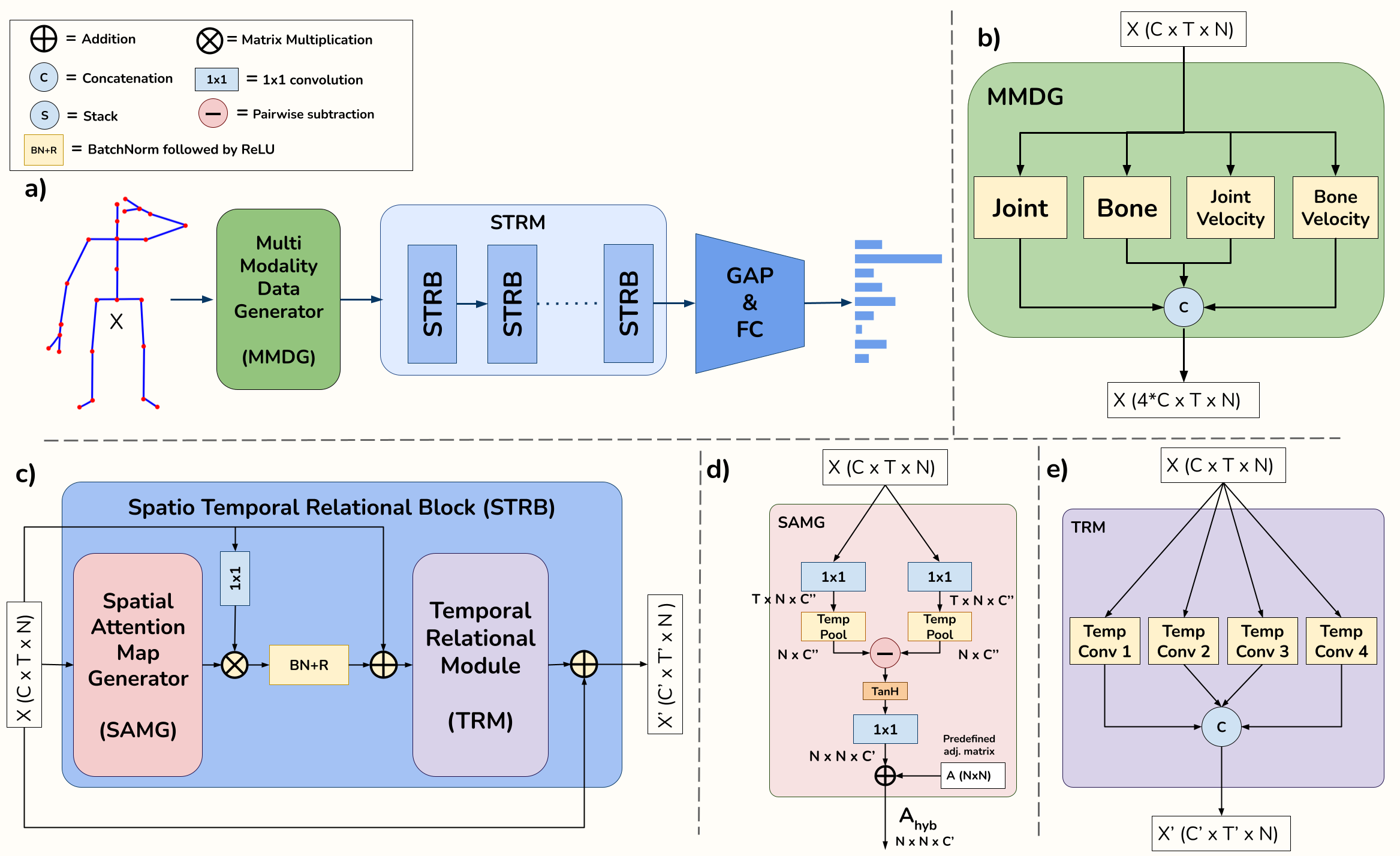}
  \caption{(a) Overall Architecture of one stream of the proposed architecture. The input skeleton is passed through Multi modality data generator (MMDG), which generates joint, bone, joint velocity and bone velocity data from input and concatenates each modality data into channel dimension as shown in (b). This multi-modal data is processed via Spatio Temporal Relational Module (STRM) followed by global average pooling and FC. (c) Spatio Temporal Relational Block (STRB), where input data is passed through Spatial Attention Map Generator (SAMG) for spatial relation modeling, followed by Temporal Relational Module. As shown in (a) multiple STRB stacked together make the STRM. (d) Spatial Attention Map Generator (SAMG), dynamically models adjacency matrix ($A_{hyb}$)to model spatial relations between joints. Predefined adjacency matrix (A) is used for regularization.  (e) Temporal Relational Module (TRM) consists of multiple temporal convolution blocks in parallel. Output of each temporal convolution block is concatenated to generate final features.}
  \label{fig:architecture}
\end{figure*}

We first describe our approach for factorizing the input skeleton into part groups and a coarser version of the skeleton (Sec.~\ref{sec:partstream}). Subsequently, we provide the architectural details of our deep network PSUMNet which processes these part streams (Sec.~\ref{sec:psumnet}).

\subsection{Part Stream Factorization}
\label{sec:partstream}


Let $X$ ($\in \mathbb{R}^{3 \times T \times N}$) represent the $T$-frames, $N$-joint skeleton configuration of a 3D skeleton pose sequence for an action. We factorize $X$ into following three part groups -- see Fig.~\ref{fig:pipeline} (right):

\begin{enumerate}
    \item \textbf{Coarse body} ($X_b$): This is comprised of all joints in the original skeleton for NTURGB+D skeleton topology, $25$ joints in total. For the 67-joint dense skeleton topology of NTU-X~\cite{trivedi2021ntux}, this stream comprises of all the body joints but without the intermediate joints of each finger for each hand. Specifically, only $6$ joints out of $21$ finger joints are considered per hand resulting in total of $37$ joints for NTU-X.
    \item \textbf{Hands} ($X_{h}$): This contains all the finger joints in each hand and the arm joints. Note that the arms are rooted at the throat joint. For NTURGB+D dataset, the number of joints for this stream is $13$ and for NTU-X, the total number of joints is $48$. 
    \item \textbf{Legs} ($X_{l}$): This includes all the joints in each of the legs. The leg joints are rooted at the hip joint. For NTURGB+D dataset the number of joints for this stream is $9$ and for NTU-X, the total number of joints is $13$.
\end{enumerate}

As shown in Fig.~\ref{fig:pipeline} (right), the part group sub skeletons are used to train three correspodning independent streams of our proposed PSUMNet ( Sec.~\ref{sec:psumnet}). As explained previously, our hypothesis is that many of the action categories are dominated by certain part groups and hence can be classified using only a subset of the entire input skeleton. To leverage this, we perform late decision fusion by performing a weighted average of the prediction scores from each of the part streams to obtain  the final classification. Crucially, we change the number of layers in each of the streams in proportion to number of input joints. We use 10, 6 and 4 layers respectively for body, hands and legs streams. This helps restrict the total number of parameters used for the entire protocol. 

In contrast with other part based approaches, the part groups in our setting are not completely disjoint. More crucially, the part groups are defined with respect to a shared global coordinate space. Though seemingly redundant due to multiple common joints across part groups, this design choice actually enables global motion information to propagate to the corresponding groups. Another significant advantage of such part stream approach is the better scalability to much denser skeleton datasets such as NTU-X~\cite{trivedi2021ntux} and to sparser  datasets such as SHREC\cite{de20173d}.

\subsection{PSUMNet}
\label{sec:psumnet}

In what follows, we explain the architecture of a single part stream of PSUMNet  (e.g. $X = X_b$) since the architecture is shared across the part streams. An overview of PSUMNet's single stream architecture can be seen in Fig.~\ref{fig:architecture} (a). First, the input skeleton $X$ is passed through Multi Modality Data Generator (MMDG) to create a rich modality aware  representation. This feature representation is processed by Spatio-Temporal Relation Module (STRM). Global average pooling ($\mathit{GAP}$) of the processed result is transformed via fully connected layers ($\mathit{FC}$) to obtain the per-layer prediction for the single part stream.

Next, we provide details for various modules included in our architecture.

\subsection{Multi Modality Data Generator (MMDG)} 
\label{sec:mmdg}

As shown in Fig.~\ref{fig:architecture} (b), this module processes the raw skeleton data and generates the corresponding multi modality data, i.e. joint, bone, joint-velocity and bone-velocity. The joint modality is the raw skeleton data represented by $X = \{x \in \mathbb{R}^{C \times T \times N}\}$, where $C$, $T$ and $N$ are channels, time steps and joints. The bone modality data is obtained using the following equation:

\begin{equation}
\centering
X_{bone} = \{x[:, \, :, \, i] \,-\, x[:, \,:, \,i_{nei}]\, |\, i = 1,2, ...,N\}
\end{equation}
where $i_{nei}$ denotes neighboring joint of $i$ based on predefined adjacency matrix. Next we create joint-velocity and bone-velocity modality data using following equations:

\begin{equation}
\centering
X_{joint-vel} = \{x[:, t+1, :] - x[:, t, :] \, |\, t = 1,2, ...,T ,\, x \in X_{joint}\}
\end{equation}
\begin{equation}
\centering
X_{bone-vel} = \{x[:, t+1, :] - x[:, t, :] \, |\, t = 1,2, ...,T ,\, x \in X_{bone}\}
\end{equation}

Finally, we concatenate all these four modality data into channel dimension to generate $X \in \mathbb{R}^{4C \times T \times N}$ which is fed as input to the network. Concatenating the modality data helps model the inter-modality relations in a more direct manner.

\subsection{Spatio Temporal Relational Module (STRM)}
\label{sec:strm}

The modality aware representation obtained from MMDG is processed by the Spatial Temporal Relational Module (STRM) as shown in Fig.~\ref{fig:architecture} (a). STRM consists of multiple Spatio Temporal Relational Blocks (STRB) stacked one after another. The architecture of a single STRB is shown in Fig.~\ref{fig:architecture} (c). Each STRB block contains a Spatial Attention Map Generator (SAMG) to dynamically model different spatial relations between joints followed by Temporal Relational Module (TRM) to model temporal relations between joints. 

\noindent \textbf{Spatial Attention Map Generator (SAMG): } We dynamically model an Spatial Attention Map for the spatial graph convolutions~\cite{chen2021channel,shi2019skeleton}. As shown in  Fig.~\ref{fig:architecture} (d), we pass the input skeleton through two parallel branches, each consisting a $1\times1$ convolution and a temporal pooling block. We  pair-wise subtract outputs from the parallel branches to model the Attention Map. We add the predefined adjacency matrix $A$ as a regularization to the Attention Map to generate the final hybrid adjacency matrix $A_{hyb}$, i.e.

\begin{equation}
\centering
A_{hyb} = \alpha M(X_{in}) + A
\end{equation}
where $\alpha$ is a learnable parameter and $A$ is the predefined adjacency matrix. $M$ is defined as:

\begin{equation}
\centering
M (X_{i}) = \sigma (\textit{TP} (\phi (X_{in})) - \textit{TP} (\psi (X_{in})) )
\end{equation}
where $\sigma$, $\phi$ and $\psi$ are 1x1 convolutions, TP is temporal pooling.

Once we obtain this adjacency matrix $A_{hyb}$, we pass the original input through a $1\times1$ convolution and multiply the results with the dynamic adjacency matrix to characterize the spatial relations between the joints as follows: 

\begin{equation}
\centering
X_{out} =  A_{hyb} \bigotimes ( \theta (X_{in}))
\end{equation}
where $\theta$ is 1x1 convolution block. $\bigotimes$ is matrix multiplication operation.

\noindent \textbf{Temporal Relation Module (TRM):} We use multiple parallel convolution blocks to model the temporal relation between the joints of the input skeleton as shown in Fig.~\ref{fig:architecture} (e). Each temporal convolution block is a standard 2D convolution with varying kernel sizes in temporal dimension and with dilation. This helps model temporal relations at multiple scales. The outputs from each of the temporal convolution blocks are concatenated. The result is processed by GAP and FC layers and mapped to a prediction (softmax) layer as mentioned previously.  

Since each part group (body, hands, legs) contains significantly different number of joints, we adjust the number of STRBs and depth of the network for each stream accordingly as shown in Fig.~\ref{fig:pipeline} (Right). This design choice provides two advantages. \textit{First,} it reduces the total number of parameters by 50\%-80\%. \textit{Second,} adjusting the depth of the network in proportion to the joint count enables richer dedicated representations for actions whose dynamics are confined to the corresponding part groups, resulting in better performance overall.


\section{Experiments}
\label{sec:experiments}

\subsection{Datasets}
\label{sec:datasets}

\textbf{NTURGB+D\cite{Shahroudy_2016_CVPR}} is a large scale skeleton action recognition dataset with 60 different actions performed by 40 different subjects. The dataset contains 25 joints human skeleton captured using Microsoft Kinect V2 cameras. There are a total of 56,880 action sequences. There are two evaluation protocols for this dataset - First, Cross Subject (XSub) split where action performed by 20 subjects falls into training set and rest into the test set. Second, Cross View (XView) protocol where actions captured via camera ID 2 and 3 are used as training set and actions captured via camera ID 1 are used as test set.
\newline
\textbf{NTURGB+D 120\cite{Liu_2019_NTURGBD120}} is an extension of NTURGB+D dataset with additional 60 action categories and a total of 113,945 action samples. The actions are performed by a total of 106 subjects. There are two evaluation protocols for this dataset - First, Cross Subject (XSub) split where action performed by 53 subjects falls into training set and rest into the test set. Second, Cross Setup (XSet) protocol where actions even setup IDs are used as training set and rest as test set.

\noindent\textbf{NTU60-X\cite{trivedi2021ntux}} is a RGB derived skeleton dataset for the sequences of the original NTURGB+D dataset. The skeleton provided in this dataset is much denser and contains 67 joints. There are total of 56,148 action samples and the evaluation protocols are same as the NTURGB+D dataset.
\newline
\noindent\textbf{NTU120-X\cite{trivedi2021ntux}} is the extension of NTU60-x dataset and corresponds to the action sequences provided by NTURGB+D 120 dataset. There are total of 113,821 samples in this dataset and the evaluation protocols are same as the NTURGB+D 120 dataset. Following \cite{trivedi2021ntux}, we evaluate our model on only Cross Subject protocol of NTU60-X and NTU120-X datasets.
\newline
\noindent\textbf{SHREC\cite{de20173d}} is a 3d skeleton based hand gesture recognition dataset. There are a total of 2800 samples with 1960 samples in train set and 840 samples in test set. Each samples has 20-50 frames and gestures are performed by 28 participants ones using only one finger and ones using the whole hand. There are 14 gestures and 28 gestures splits provided by the creators and we report results on both of these splits.

\subsection{Implementation and Optimization details}

As shown in Fig.~\ref{fig:pipeline} (right), the input skeleton to each of the part stream contains different number of joints. For NTURGB+D dataset, the body stream has input skeleton with a total of $25$ joints, hands stream has the input skeleton with a total of $13$ joints and legs stream with a total of $9$ joints. Within the PSUMNet architecture, we use $10$ STRBs for the body stream, $6$ STRBs for the hands stream and $4$ STRBs to process the legs stream.

We implement PSUMNet using the Pytorch deep learning framework. We use SGD optimizer with $0.1$ as the base learning rate and a weight decay of $0.0005$. All the models are trained on 4 1080Ti 12 GB GPU systems. For training of 25 joints datasets-NTU60 and NTU120, we use a batch size of 200. For 67 joints datasets-NTU60-X and NTU120-X, due to much denser skeleton, smaller batch size of 65 is used. 

\subsection{Results}
\label{sec:results}

\begin{table*}[!t]
  \resizebox{\textwidth}{!} 
    {%
    \centering
  \begin{tabular}{l|l|cc|cc|cc}
    \toprule
    & & & & \multicolumn{2}{c|}{NTU60} & \multicolumn{2}{c}{NTU120}\\
    Type & Model & Params. (M) $^*$ & FLOPs (G) $^*$ &  XSub & XView & Xsub & XSet\\
    \midrule
    \multirow{2}{4em}{CNN based} & VA-Fusion \cite{zhang2019view} & 24.6 & - & 89.4 & 95.0 & - & - \\
    & TaCNN+ \cite{xu2021topology} & $\,\,$4.4 & $\,\,\,$1.0 & 90.7 & 95.1 & 86.7 & 87.3\\
    \midrule
    \multirow{10}{4em}{GCN based} & ST-GCN \cite{stgcn2018aaai} & $\,\,$3.1 & $\,$16.3 & 81.5 & 88.3 & 70.7 & 73.2 \\
    & RA-GCN \cite{song2020richly} & $\,\,$6.2 & $\,$32.8  & 87.3 & 93.6 & 81.1 & 82.7\\
    & 2s-AGCN \cite{shi2019two} & $\,\,$6.9 & $\,$37.3 & 88.5 & 95.1 & 82.9 & 84.9\\
    & PA-ResGCN\cite{song2020stronger} & $\,\,$3.6 & $\,$18.5 & 90.9 & 96.0 & 87.3 & 88.3\\
    & DDGCN\cite{korban2020ddgcn} & - & - & 91.1 & $\mathbf{97.1}$ & - & -\\
    & DGNN \cite{shi2019skeleton} & 26.2 & - & 89.9 & 95.0 & - & - \\
    & MS-G3D\cite{liu2020disentangling} & $\,\,$6.4 & $\,$48.8 & 91.5 & 96.2 & 86.9 & 88.4\\
    & 4s-ShiftGCN\cite{cheng2020shiftgcn} & $\,\,$2.8 & $\,$10.0 & 90.7 & 96.5 & 85.9 & 87.6\\
    & DC-GCN+ADG \cite{cheng2020decoupling} & $\,\,$4.9 & $\,$25.7 & 90.8 & 96.6 & 86.5 & 88.1\\
    & DualHead-Net \cite{chen2021learning} & 12.0 & - & 92.0 & 96.6 & 88.2 & 89.3\\
    & CTR-GCN \cite{chen2021channel} & $\,\,$5.6 & $\,\,\,$7.6 & 92.4 & $96.8$ & 88.9 & 90.6\\
    \midrule
    \multirow{2}{4em}{Attention based} & DSTA-Net\cite{dstanet_accv2020} & 14.0 & $\,$64.7 & 91.5 & 96.4 & 86.6 & 89.0\\
    & ST-TR \cite{plizzari2021skeleton} & 12.1 & 259.4 & 89.9 & 96.1 & 82.7 & 84.7\\
    & 4s-MST-GCN \cite{chen2021multi} & 12.0 & - & 91.5 & 96.6 & 87.5 & 88.8\\
   
    \midrule
    & \textbf{PSUMNet(Ours)} & $\,\,2.8$ & $\,\,\,$2.7 & $\mathbf{92.9}$ & 96.7 & $\mathbf{89.4}$ & $\mathbf{90.6}$ \\
    
  \bottomrule
\end{tabular}
}
\caption{\label{tab:ntu_results}Comparison with state of the art approaches for NTURGB+D and NTURGB+D 120 dataset. Model parameters are in millions ($\times10^6$) and FLOPs are in billions ($\times10^9$). $^*$: These numbers are cumulative over all the streams used by respective models as per their training protocol.}
\end{table*}

\begin{table}[!t]
  \resizebox{0.75\linewidth}{!} 
    {%
  \begin{tabular}{l|c|cc}
    \toprule
    Model & Params. (M) &  NTU60 & NTU120\\
    \midrule
    PA-ResGCN\cite{song2020stronger} & 3.6 & 90.9 & 87.3 \\
    MS-G3D (Joint)\cite{liu2020disentangling} & 3.2 & 89.4 & 84.5 \\
    1s-ShiftGCN (Joint)\cite{cheng2020shiftgcn} & 0.8 & 87.8 & 80.9\\
    DSTA-Net (Bone)\cite{dstanet_accv2020} & 3.5 & 88.4 & 84.4\\
    DualHead-Net (Bone)\cite{chen2021learning} & 3.0 & 90.7 & 86.7 \\
    CTR-GCN (Bone)\cite{chen2021channel} & 1.4 & 90.6 & 85.7\\
    TaCNN+ (Joint)\cite{xu2021topology} & 1.1 & 89.6 & 82.6\\
    MST-GCN (Bone)\cite{chen2021multi} & 3.0 & 89.5 & 84.8\\
    \midrule
    \textbf{PSUMNet(Ours)} (Body) & 1.4 & $\mathbf{91.9}$ & $\mathbf{88.1}$\\
    
  \bottomrule
  
\end{tabular}
}
\caption{\label{tab:results_single_stream}Comparison of only body stream of PSUMNet with the best performing modality (i.e only joint, only bone) of state of the art approaches for NTURGB+D 60 and 120 dataset on Cross Subject protocol.}
\end{table}

\begin{table}[!t]
  \resizebox{0.75\linewidth}{!} 
    {%
  \begin{tabular}{l|c|cc}
    \toprule
    Model & Params. (M) &  NTU60-X & NTU120-X\\
    \midrule
    PA-ResGCN\cite{song2020stronger} & $\,\,$3.6 & 91.6 & 86.4 \\
    MS-G3D\cite{liu2020disentangling} & $\,\,$6.4 & 91.8 & 87.1 \\
    4s-ShiftGCN\cite{cheng2020shiftgcn} & $\,\,$2.8 & 91.8 & 86.2\\
    DSTA-Net\cite{dstanet_accv2020} & 14.0 & 93.5 & 87.8\\
    CTR-GCN \cite{chen2021channel} & $\,\,$5.6 & 93.9 & 88.3\\
    \midrule
    \textbf{PSUMNet(Ours)} & $\,\,3.2$ & $\mathbf{94.7}$ & $\mathbf{89.1}$\\
  \bottomrule
\end{tabular}
}
\caption{\label{tab:ntux_results}Comparison with state of the art approaches for dense skeleton datasets NTU60-X and NTU120-X.}
\end{table}

Tab.~\ref{tab:ntu_results} compares the performance of proposed PSUMNet with other approaches on Cross Subject (XSub) and Cross View (XView) splits of NTURGB+D dataset \cite{Shahroudy_2016_CVPR} and Cross subject (XSub) and Cross Setup (Xset) splits of the NTURGB+D 120 dataset\cite{Liu_2019_NTURGBD120}. As can be seen from the Params. column in Tab.~\ref{tab:ntu_results}, PSUMNet uses the least number of parameters compared to other methods and achieves better or very comparable results across different splits of the datasets. For the harder Cross Subject split of both NTURGB+D and NTURGB+D 120, PSUMNet achieves state of the art performance compared to other approaches which use 100\%-400\% more parameters. This shows the superiority of PSUMNet both in terms of performance and efficiency - also see Fig.~\ref{fig:teaser}.

We also compare the performance of only body stream of PSUMNet with single stream (i.e only joint, only bone) performance of other approaches in Tab.~\ref{tab:results_single_stream} for Xsub split of NTURGB+D and NTURGB+D 120 datasets. As can  be seen, PSUMNet outperforms other approaches by a margin of 1-2\% for NTURGB+D and by 2-3\% for NTURGB+D 120 using almost the same or lesser number of parameters. This also supports our hypothesis that part stream based unified modality approach is much more efficient compared to conventional independent modality streams approach.

\begin{table}[!t]
  \resizebox{0.75\linewidth}{!} 
    {%
  \begin{tabular}{l|c|cc}
    \toprule
    Model & Params. (M) &  14 gestures & 28 gestures\\
    \midrule
    Key-Frame CNN \cite{de20173d} & $\,\,$7.9 & 82.9 & 71.9 \\
    CNN+LSTM \cite{nunez2018convolutional} & $\,\,$8.0 & 89.8 & 86.3 \\
    Parallel CNN\cite{devineau2018convolutional} & 13.8 & 91.3 & 84.4\\
    STA-Res TCN \cite{hou2018spatial} & $\,\,$6.0 & 93.6 & 90.7\\
    DDNet \cite{yang2019make} & $\,\,$1.8 & 94.6 & 91.9\\
    DSTANet \cite{dstanet_accv2020} & 14.0 & $\mathbf{97.0}$ & $\mathbf{93.9}$ \\
    \midrule
    \textbf{PSUMNet(Ours)} & $\mathbf{\,\,0.9}$ & 95.5 & 93.1\\
  \bottomrule
\end{tabular}
}
\caption{\label{tab:shrec_results}Comparison with state of the art approaches for SHREC skeleton hand gesture recognition dataset.}
  
\end{table}

Trivedi et al.\cite{trivedi2021ntux} introduced NTU60-X and NTU120-X, extensions of existing NTURGB+D and NTURGB+D 120 datasets with 67 joint dense skeletons containing fine-grained finger joints within the full body pose tree. Handling such large number of joints while keeping the total parameters of the model in bounds is a difficult task. However, as shown in Tab.~\ref{tab:ntux_results}, PSUMNet achieves state of the art performance for both NTU60-X and NTU120-X datasets. Total parameters increase by a small margin for PSUMNet to handle the additional joints, yet it is worth noting that other competing approaches use 100\%-400\% more parameters as compared to PSUMNet. This shows the benefit of using part based streams approach for dense skeleton representation as well.

To further explore the generalization capability of our proposed method, we evaluate performance of PSUMNet for skeleton based hand gestures recognition dataset, SHREC\cite{de20173d}. Taking advantage of part based stream approach, we  train only the hands stream of PSUMNet. As shown in Tab.~\ref{tab:shrec_results}, PSUMNet achieves comparable results to existing state of the art method (DSTANet\cite{dstanet_accv2020}) which uses 1400\% more parameters. PSUMNet  outperforms the second best approach (DDNet\cite{yang2019make}) which uses 100\% more parameters.

Overall, Tab.~\ref{tab:ntu_results}, \ref{tab:ntux_results} and \ref{tab:shrec_results} comprehensively show that proposed PSUMNet achieves state of the art performance, generalizes easily across a range of action datasets and uses a significantly smaller number of parameters compared to other methods.

\subsection{Analysis}
\label{sec:discussion}

\begin{figure*}[!t]
  \centering
  \includegraphics[width=0.8\textwidth]{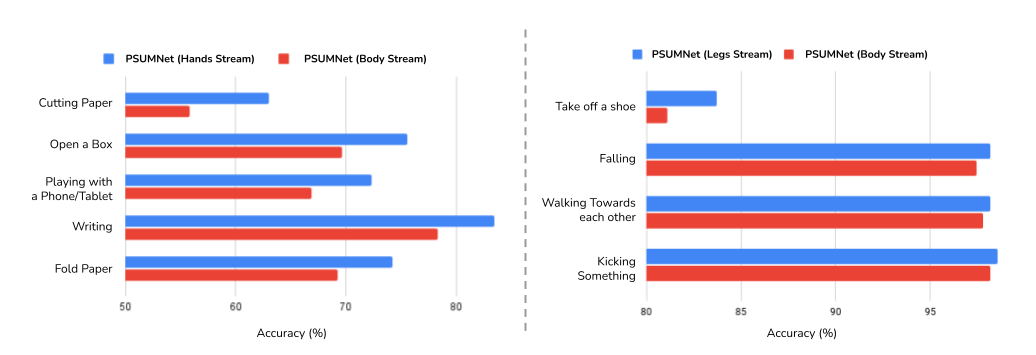}
  \caption{Comparing per class accuracy after training PSUMNet using only Hands stream and only body stream for NTU120-X dataset (Left) and only Legs stream with only body stream for NTU60-X datset (Right). On observing the class labels we can see that all the actions in the left plot are dominated by hand joints movements and all the actions in the right plot are dominated by leg joints movement and hence streams corresponding to these parts are able to classify these classes better which is in line with our hypothesis}
  \label{fig:hands_body_120x}
\end{figure*}

\begin{figure}[!t]
  \centering
  \includegraphics[width=0.4\linewidth]{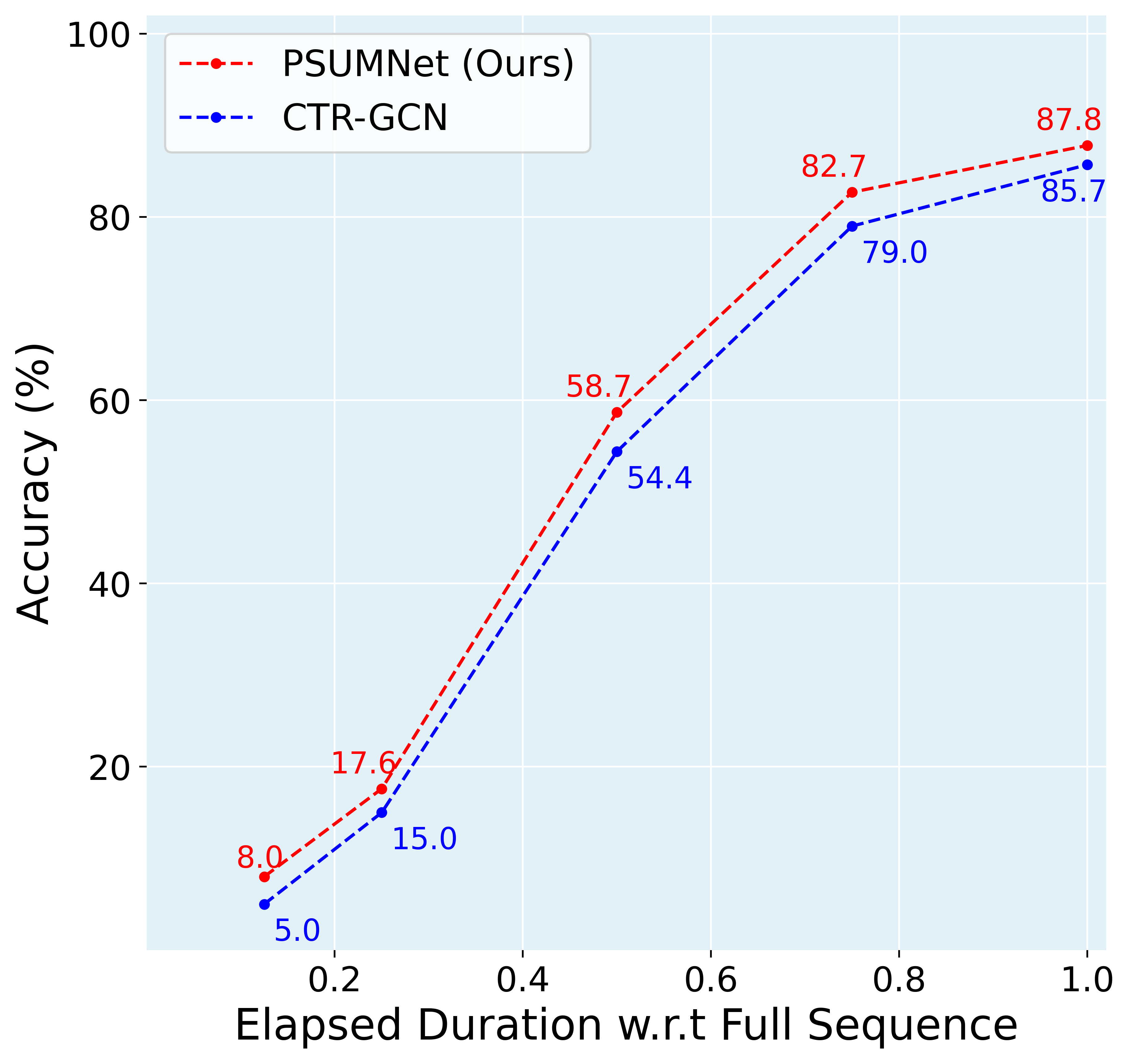}
  \caption{Comparing PSUMNet with current state of the art method, CTR-GCN on partially observed sequences for NTURGB+D 120 (XSub) dataset. Annotated numbers for each line plot denote accuracy of both models on partial sequences.} \label{fig:temp_analysis}
\end{figure}

As explained in Sec.~\ref{sec:partstream}, we train PSUMNet using three part streams namely body, hands and legs streams and report the ensembled results from all the three streams. To understand the advantage of the proposed part stream approach, we compare stream wise per class accuracy for NTU120-X and NTU60-X of PSUMNet. Fig.~\ref{fig:hands_body_120x} (left) depicts the per class comparison setting for per class accuracy comparison between the `only hands stream' and `only body stream' setting of PSUMNet for NTU120-X dataset. The classes shown correspond to those with largest (positive) gain in per class accuracy while using only hand stream. Upon observing the action labels of these classes, (``Cutting Paper", ``Writing", ``Folding Paper"), it is evident that these classes are dominated by hand joints movements and hence are better classified using only a subset of input skeleton which has dedicated representations for hand joints as opposed to using entire skeleton in a monolithic fashion. 

Similarly, we also compare the per class accuracy while using only legs stream against only body stream of PSUMNet for NTU60-X dataset as shown in Fig.~\ref{fig:hands_body_120x} (right). In this case too, the class labels with highest positive gain while using only legs stream are dominated correspond to expected classes such as ``Walking", ``Kicking". 

The above results can also be appreciated better by studying the number of parameters in each of the part based stream. The body stream in PSUMNet has $1.4M$ parameters, Hands stream has $0.9M$ and legs stream has $0.5M$ parameters. Hence, hands stream while using only $65\%$ of the total parameters of the body stream and legs stream while using only $35\%$ of the body stream parameters can identify those classes better which are dominated by joints corresponding to each part stream. 

\noindent \textit{Early action recognition:} In the experiments so far, evaluation was conducted on the full action sequence. In other words, the predicted label is known only after all the frames are provided to the network. However, there can be anticipatory scenarios where we may wish to know the predicted action label without waiting for the entire action to finish. To examine the performance in such scenarios, we create test sequences whose length is determined in terms of a fraction of the total sequence length. We study the trends in accuracy as the \% of total sequence length is steadily increased. For comparison, we study PSUMNet with the state of the art network, CTR-GCN~\cite{chen2021channel}. As can be seen in Fig.~\ref{fig:temp_analysis}, PSUMNet consistently outperforms CTR-GCN for partially observed sequences, indicating its suitability for early action recognition.

\subsection{Ablations}
\label{sec:ablation}

To understand the contribution of each part stream in PSUMNet, we provide individual stream wise performance of PSUMNet on NTU60 and NTU120 datasets Cross Subject splits as ablations in Tab.~\ref{tab:ablation_table}.

At a single stream level, the body stream achieves higher accuracy compared to hands and legs stream. This is expected since the body stream includes a coarse version of all the joints. However, as mentioned previously (Sec.~\ref{sec:discussion}), hands and legs streams classify actions dominated by respective joints better. Therefore, accuracies of Body+Hands (row 4 in Tab.~\ref{tab:ablation_table}) and Body+Legs (row 5) variants are higher than  only the body stream. Legs stream achieves lower accuracy as compared to body and hands stream because there are only a small subset of action categories which are dominated by leg joints movements. However, as with hands stream, legs stream benefits classes which involve significant leg joints movements.

\begin{table}[!t]
  \resizebox{0.7\linewidth}{!} 
    {%
  \begin{tabular}{l|l|c|cc}
    \toprule
    Type & Stream & Params. (M) & NTU60 & NTU120\\
    \midrule
    \multirow{6}{3em}{Part Streams} 
    & Body & 1.4 & 91.9 & 88.1\\
    & Hands & 0.9 & 90.3 & 85.8\\
    & Legs & 0.5 & 60.4 & 50.6\\
    & Body + Hands & 2.3 & 92.4 & 89.0\\
    & Body + Legs & 1.9 & 92.1 & 87.9\\
    & Hands + Legs & 1.4 & 90.9 & 86.5\\
    \midrule
    Disjoint Parts & Body + Hands + Legs 
    & 2.8 & 89.6 & 86.1\\
    \midrule
    \multirow{5}{3em}{Modalities in PSUMNet} 
    & Joint & 2.8 & 90.3 & 86.1\\
    & Bone & 2.8 & 90.1 & 87.6\\
    & Joint-Vel & 2.8 & 88.5 & 82.7\\
    & Bone-Vel & 2.8 & 87.6 & 83.2\\
    & Joint + Bone & 2.8 & 91.4 & 88.6\\
    \midrule
    \multicolumn{2}{c|}{\textbf{PSUMNet}} & 2.8 & $\mathbf{92.9}$ & $\mathbf{89.4}$\\
  \bottomrule
 
\end{tabular}
}
 \caption{\label{tab:ablation_table} Indivdual streams performance on NTURGB+D and NTURGB+D 120 Cross Subject dataset.}
  
\end{table}


Our proposed part groups factorization registers each group's sub-skeleton in a global frame of reference (see Fig.~\ref{fig:pipeline}). Further, all the part groups are not disjoint and have overlapping joints to better propagate global motion information through the network (Sec.~\ref{sec:partstream}). To justify our choice of globally registered part groups, we perform an ablation with a different part grouping strategy, with each part group being disjoint and in a local frame of reference. Specifically, the ablation setup for body stream includes on $9$ torso joints (including shoulders and hips joints), hands stream includes only $12$ joints and legs stream includes only $8$ joints. It is important to notice here that unlike our original strategy, both legs and hands in corresponding part stream are not connected. As expected, such strategy fails to capture global motion information unlike our proposed method (c.f. `Disjoint parts' row and last row in Tab.~\ref{tab:ablation_table}).

To further investigate contribution of each data modality in our proposed unified modality method, we provide ablation studies with PSUMNet trained on single and two modalities instead of four (c.f. `Modalities in PSUMNet' rows and last row in Tab.~\ref{tab:ablation_table}). We notice that PSUMNet benefits most by joint and bone modalities compared to velocity modalities. However, the best performance is obtained by utilizing all the modalities.

\section{Conclusion}
\label{sec:conclusion}

In this work, we present Part Streams Unified Modality Network PSUMNet to efficiently tackle the challenging task of scalable pose-based action recognition. PSUMNet uses part based streams and avoids treating the input skeleton in monolithic fashion as done by contemporary approaches. This choice enables richer  and dedicated representations especially for actions dominated by a small subset of localized joints (hands, legs). The unified modality approach introduced in this work enables efficient utilization of the inter-modality correlations. Overall, the design choices provide two key benefits -- (1) they help attain state of the art performance using significantly smaller number of parameters compared to existing methods (2) they allow PSUMNet to easily scale to both sparse and dense skeleton action datasets in distinct domains (full body, hands) while maintaining high performance. PSUMNet is an attractive choice for pose-based action recognition especially in real world deployment scenarios involving compute restricted embedded and edge devices.

%
%
\bibliographystyle{splncs04}
\bibliography{main}
\end{document}